# ALPET: Active Few-shot Learning for Citation Worthiness Detection in Low-Resource Wikipedia Languages

Aida Halitaj, Arkaitz Zubiaga

School of Electronic Engineering and Computer Science
Queen Mary University of London
London, United Kingdom
{a.halitaj, a.zubiaga}@qmul.ac.uk

**Abstract**

Citation Worthiness Detection (CWD) determines which sentences, within an article, should be backed up with a citation to validate the information it provides. This study, introduces ALPET, a framework combining Active Learning (AL) and Pattern-Exploiting Training (PET), to enhance CWD for languages with limited data resources. Applied to Catalan, Basque, and Albanian Wikipedia datasets, ALPET outperforms the existing CCW baseline while reducing the amount of labeled data in some cases above 80%. ALPET's performance plateaus after 300 labeled samples, showing it suitability for low-resource scenarios where large, labeled datasets are not common. While specific active learning query strategies, like those employing K-Means clustering, can offer advantages, their effectiveness is not universal and often yields marginal gains over random sampling, particularly with smaller datasets. This suggests that random sampling, despite its simplicity, remains a strong baseline for CWD in constraint resource environments. Overall, ALPET's ability to achieve high performance with fewer labeled samples makes it a promising tool for enhancing the verifiability of online content in low-resource language settings.

*Keywords:* Low-Resource Languages, Wikipedia, Few-Shot Learning, Active Learning, Fact Checking, Citation Worthiness Detection

## 1. Introduction

The rise of misinformation in the digital age has made automated fact-checking an essential tool in ensuring the reliability of information [1, 2]. Automated fact-checking are complex systems including tasks that involve the identification of information worthy of checking, linking it with evidence and the subsequent verification step [3]. While substantial progress has been made in fact-checking systems for major languages, low-resource languages remain largely underexplored and are recently receiving more attention [4, 5].

Misinformation has a global impact [6], so contributing towards advancement of fact-checking systems for low-resource languages is necessary to maintain information integrity worldwide. However, in doing so, the scarcity of labeled datasets for low-resource languages poses a significant challenge [7]. Many of the existing methods for developing automated fact-checking systems rely on extensive labeled data, which is often unavailable for low-resource languages. Furthermore, in high-resource languages, like English, fact-checking systems have the advantage of focusing on specific domains, such as political discourse, which generates abundant data for verification [8]. In contrast, low-resource languages often lack this advantage, as the volume of digital information available in specialized topics is more limited, affecting the ability to build effective systems within a single domain of a language [7, 9]; as a result, we cannot limit ourselves to one specific domain for a low-resource language.

In this research we address this challenge by using Wikipedia data, which spans a wide range of topics and domains. The goal is to develop a method for citation worthiness detection (CWD), where the task is to determine whether a sentence needs to be accompanied with a citation to back it up; an ability to detect sentences needing citation in turn supports the integrity of Wikipedia. Research in CWD has been limited to date, with the majority of efforts focused on English [10]; when it comes to low-resource languages, the only such effort to date was with CCW [4], where we introduced a context-aware model that leverages surrounding sentences and topic categories to help determine if a sentence should be accompanied with a citation.

Furthering the limited research of CWD in low-resource language, here we are the first to propose an active learning strategy specifically tailored to address data scarcity in the task. We introduce a novel method called ALPET (Active Learning with Pattern Exploiting Training), which integrates active learning (AL) strategies with few-shot learning (FSL) techniques using pre-trained language models (PLMs). This approach is designed to maximize performance with minimal labeled data, addressing the limitations of existing methods for low-resource languages. ALPET strategically selects the most informative samples for labeling, thus reducing the need for extensive labeled datasets, which are often unavailable for low-resource languages. While few-shot learning (FSL) enables models to generalize from a small number of examples [11], AL further optimizes this process by focusing on high-value samples, reducing the cost and time associated with manual labeling. Despite the promise of both AL and PET, their integration for low-resource languages in CWD tasks has not been explored until now.

The goal of ALPET is to investigate whether AL strategies, when combined with PET, could improve model performance as opposed to the existing models like CCW [4]. Furthermore, in this research we investigate if ALPET can select more informative samples than random selection. While we initially hypothesised that this approach would consistently outperform random sampling, the results indicate that the effectiveness of AL strategies is more nuanced and context-dependent. Despite this, the study provides valuable insights into the challenges of applying AL and FSL methods in low-resource language settings, offering a foundation for future work in optimizing data selection and model performance in these environments.

Our study makes the following contributions:

1. We introduce ALPET, a novel active few-shot learning method that integrates active learning (AL) strategies with few-shot learning (FSL) techniques using pre-trained language models (PLMs) through Pattern Exploit Training (PET). ALPET leverages Wikipedia data to address the data scarcity issue in low-resource languages by strategically selecting the most informative samples for labelling, reducing the dependency on large, labelled datasets.

2. We provide a comprehensive analysis of multiple variants of our proposed ALPET framework, including with various AL query strategies for CWD, comparing their performance against random sampling in a few-shot learning environment. The analysis focuses on the effectiveness of these strategies in identifying informative data points for labelling and its suitability for CWD in low-resource settings, with a particular emphasis on K-Means clustering techniques.

3. The research quantifies the reduction in labelled data achieved by ALPET compared to existing methods like CCW while maintaining comparable performance. This reduction in manual annotation not only improves efficiency but also significantly reduces development costs, making it particularly valuable in low-resource contexts.

4. Within the largely unexplored realm of CWD for low-resource languages, ours is the first effort to incorporate and investigate active learning, in turn comparing with the state-of-the-art method, CCW, and providing a comprehensive analysis of the potential benefits of the proposed strategy.

These contributions offer a deeper understanding of the challenges and potential solutions for CWD in low-resource languages and pave the way for further advancements in the field by establishing a foundation for building effective CWD systems for under-resourced languages.

## 2. Research Objectives and Hypotheses

The scarcity of labeled data in low-resource languages presents a major challenge for developing automated fact-checking systems. In this research, we aim to address this issue by exploring how the combination of AL and PET can enhance citation check-worthiness detection for low-resource languages, which in our case we investigate with Albanian, Basque, and Catalan.

The specific objectives (O) of this research, and the hypotheses (H) we set forth in line with the objectives, are:

- **O1:** To investigate whether ALPET can achieve comparable or superior citation worthiness detection performance in low-resource languages (Albanian, Basque, and Catalan) compared to the CCW baseline model, while utilising fewer labelled examples. This objective directly addresses hypothesis H1, which posits that ALPET will outperform CCW in data efficiency and F1 score in low-resource languages.

  **H1:** ALPET outperforms the CCW baseline model in terms of data efficiency (fewer labeled examples) and performance (F1 Score) in low-resource languages while using the same AL query strategies.

- **O2:** To determine the optimal number of labelled examples required by ALPET to achieve peak performance in citation worthiness detection for low-resource languages, analysing whether the model's performance plateaus after a certain number of labelled samples. This objective is directly related to hypothesis H2, which suggests that ALPET's performance will plateau after a certain number of samples, highlighting its effectiveness with small datasets.

  **H2:** ALPET's performance improves with increasing labeled data but it may reach a plateau at a certain point, suggesting that the method is effective with small sized datasets in low-resource settings.

- **O3:** To quantify the reduction in labelled data achieved by ALPET compared to the CCW baseline model while maintaining comparable citation worthiness detection performance in low-resource languages, assessing the robustness of ALPET's performance with reduced training data. This objective is linked to hypothesis H3, which predicts that ALPET will achieve competitive performance with an average of 58-72% fewer labelled examples than CCW, demonstrating its robustness in low-resource settings.

  **H3:** ALPET can match the performance of CCW with far fewer labeled examples in low-resource languages. Its performance stays stable even as the number of labeled examples decreases, showing its robustness in these settings.

- **O4:** To compare the performance of various active learning query strategies against random sampling in selecting informative data points for citation worthiness detection in low-resource languages, evaluating their effectiveness based on the F1 Score. This objective directly addresses hypothesis H4, which states that active learning query strategies generally achieve higher F1 scores than random sampling in low-resource languages.

  **H4:** Active learning query strategies generally achieve higher F1 scores than random sampling in low-resource language datasets.

## 3. Background and Related Work

### 3.1. Citation Worthiness Detection (CWD)

Citation worthiness detection, often referred to as the "citation needed" task in the literature, involves identifying whether a sentence within a given corpus requires a citation [12, 10]. This task is particularly crucial on collaborative platforms like Wikipedia, where maintaining information credibility is essential. By ensuring that unsourced statements are flagged to be properly supported by reliable references, CWD helps prevent the spread of misinformation. CWD in Wikipedia simplifies and speeds up the process of verifiability policy[1] by prioritizing unsourced sentences to be reviewed by editors. This process is vital for maintaining academic and public trust in Wikipedia, which serves as a widely-used reference for both educational and general purposes. It is also critical in fact-checking systems to evaluate the veracity of claims [13], and in various social media platforms to combat misinformation [14, 4].

---

[1] https://en.wikipedia.org/wiki/Wikipedia:Verifiability

CWD in low-resource languages has several challenges. One major issue is the availability of credible labeled datasets, as these languages often lack the extensive digital content needed to develop such resources. As a result, most NLP tools and pre-trained models, like BERT and GPT, are optimized for high-resource languages, leading to suboptimal performance in low-resource contexts. Specifically, considering that low-resource languages have variety of dialects that are underrepresented or not captured at all by PLMs. The scarcity of scientific research focused on low-resource languages further amplifies the problem, as most advancements in the field are designed and tested on larger languages.

Existing CWD approaches in Wikipedia are usually defined as supervised learning text classification task. The pioniring work for this task started from the assessment of Wikipedia verifiability policy [10] where they used recurrent neural networks (RNN) with GRU cells and GloVe pre-trained word embeddings to identify sentences that needed citation. However, they heavily relied on featured articles [2], citation needed tag [3], and manual annotation efforts to create the dataset.

Building upon Redi's work, another approach was proposed utilizing positive unlabeled learning where they aimed to develop an unified approach to check-worthiness detection tasks including claim detection, rumour detection and citation needed [15]. While aiming to create a unified solution for these three distinct tasks, authors also aimed to reduce manual labelling effort by asking annotators to mark only sentences that were clear-cut check-worhty and the rest to be handled through positive unlabeled method. They used transfer-learning to transfer the knowledge from one task to another and different from more traditional neural networks used in Redi's approach, in this research they used pre-trained BERT model.

Both studies [10] and [15] focused primarily on English, leveraging featured articles and {*citation needed*} tags added by active editors. These methods, however, are not applicable to low-resource languages due to the scarcity of featured articles and the absence of such tags, resulting from a lack of active editorial communities. To overcome this challenge, a recent study proposed an approach to use quality score of articles to automatically build a credible datasets for low-resource languages [4]. Unlike previous work, which relied solely on the sentence text and its section placement within an article, this study used adjacent sentences as contextual information and employed mBERT for the final classification of sentences needing citations. While their approach advances CWD in low-resource languages, it relies on substantial amount of data due to the need for contextual information from adjacent sentences; and the automated large-scale labeling process, although innovative, carries the challenge of potential inaccuracies, as it cannot fully ensure the correctness of every label. Incorporating an oracle in the labeling process through active learning could address these issues by enhancing the credibility of the dataset. Building on this foundation, we propose an approach that integrates cold-start pool-based active learning with few-shot learning using Pattern Exploit Training (PET), which reduces the data requirements for effective CWD and minimizes dependence on additional contextual information.

### 3.2. Active Learning in NLP

Active learning (AL) is a machine learning approach where the algorithm uses a querying strategy to identify the most informative data points for labeling by an oracle, to improve model's performance [16]. The goal is to overcome the labeling bottleneck of traditional passive learning systems by optimizing the model's performance with a smaller set of labeled examples, making the learning process more efficient and cost-effective. This approach is particularly valuable in scenarios with limited labeled data, such as low-resource Natural Language Processing (NLP) tasks.

AL involves two key concepts: problem scenarios and query strategies. Scenarios define the learning environment, including how data is presented and how the model interacts with it, while query strategies determine which data points to label within that scenario. Some query strategies can be applied across multiple scenarios, while others are scenario-specific. For an AL system to be effective, it is essential to match the appropriate query strategy with the right scenario. According to the existing literature some of the main AL scenarios are membership query synthesis [17], stream-based selective sampling [18], pool-based sampling [19], batch AL [20], and multi-task AL [21, 22, 23].

---

[2] `https://en.wikipedia.org/wiki/Wikipedia:Featured_articles`
[3] `https://en.wikipedia.org/wiki/Wikipedia:Citation_needed`

Each scenario offers distinct advantages and has seen different levels of application in NLP. In what follows we will briefly introduce the above mentioned AL scenarios.

1. *Membership Query Synthesis (MQS)*. It is one of the earliest AL scenarios [24] which enables the generation of artificial examples to expand datasets based on defined feature dimensions. Initially endorsed for automated oracles [25, 26] due to the challenges humans faced in interpreting synthetic examples, MQS has recently been applied to simple text classification tasks [27, 28]. Nonetheless, its use remains limited, specifically for more complex NLP tasks with unbalanced data [27, 29].

2. *Stream-Based Selective Sampling*. Also known as online AL [30], this scenario involves data arriving in a continuous stream, requiring the learner to decide in real-time whether to request the label [16]. Query strategies like uncertainty sampling, threshold-based decision, query by committee are commonly employed to make this decision effective. This scenario has been applied in several NLP tasks, including part-of-speech tagging [31], named entity recognition [32], sentiment analysis [33, 34] etc.

3. *Pool-Based Sampling*. This scenario, perhaps the most extensively researched in NLP [35], involves selecting the most informative samples from a large, static pool of unlabeled data [19]. It has been successfully applied in a wide range of NLP tasks, including text classification [19, 36, 37, 38, 39], speech recognition [40], named entity recognition [41, 42, 43], part of speech tagging [44, 45, 46], word sense disambiguation [47, 48, 49], machine translation [50, 51], language understanding[52], and prompt engineering [53]. The versatility of this approach is further enhanced by combining it with deep learning techniques like transfer-learning [39, 54], semi-supervised learning [55, 56], weak supervision [57, 58, 59], data augmentation [50, 43], and few-shot learning [60, 61, 62, 53, 63], resulting in development of various query strategies adhering to deep neural network (DNN) characteristics [64]. However, its application in more specialized tasks, such as citation worthiness detection, remains underexplored.

4. *Batch Active Learning*. Unlike traditional AL, where data points are queried one by one, batch AL queries multiple data points simultaneously to increase efficiency and reduce the number of iterations needed for model improvement [16, 65]. This approach has been applied in various NLP tasks, including text classification [66], machine translation [67, 68], rumor detection [69], and named entity recognition (NER) [70, 68]. However, ensuring diversity among selected examples remains a challenge [71, 72], particularly in complex NLP tasks with unbalanced data, where similar sentences and repetition can reduce information gain.

5. *Multi-Task Active Learning (MTAL)*. It handles multiple tasks simultaneously, improving efficiency by allowing related tasks to share data and enhancing performanc; for distinct tasks by ensuring comprehensive annotations [21, 16]. It has been applied in NLP tasks such as role labeling [73], dependency parsing [74], named entity recognition [75, 74], and natural language understanding [76]. While challenges like data scarcity and annotation complexity exist across different AL scenarios, they are magnified in MTAL due to the need to manage multiple tasks simultaneously. This adds layers of complexity and resource demands that are less noticeable in single-task scenarios, making it less suited for complex tasks like those in fact-checking.

Pool-based sampling is one of the most used AL query strategy in NLP tasks due to its effectiveness in minimizing labeling costs by selecting only the most informative data points. This approach allows modern NLP models, specifically PLMs, such as transformers (e.g., BERT [77], GPT [78], T5 [79]), to achieve high performance with limited resources [80, 81, 82, 83, 84], particularly in low-resource languages and domains [85, 52, 86, 87, 88, 89, 83]. It also facilitates rapid domain adaptation [82] and enhances human-in-the-loop systems by focusing efforts on the most impactful examples [81]. Yet, its application in more specialized tasks, such as citation worthiness detection, remains underexplored – a gap we aim to address in our research.

While AL has been used in related areas like misinformation detection [90], rumour detection [69], claim verification [62], its use in CWD in Wikipedia settings is limited. These studies primarily focus in specific domains like political fact-checking and English language, rather than the broader, domain-agnostic context of Wikipedia. The gap in the literature suggests a need for further exploration in applying AL to CWD task, particularly with an emphasis on language diversity and resource constraints.

In the context of fact-checking, AL has been combined with models like PET to improve the accuracy of claim verification [62]. However, similar integration for citation worthiness detection or claim detection (as a related task), especially within low-resource language settings, is not common. Our work seeks to address this gap by exploring different pool-based AL query strategies in conjunction with PET models to improve CWD in low-resource settings.

In the next section we explain in more details our proposed methodology.

## 4. Methodology

The proposed methodology, referred to as ALPET, integrates Active Learning (AL) and Pattern Exploit Training (PET) to create an efficient approach for data selection and model training in low-resource setting. It is structured into four key steps:

1. Data selection: the process begins with applying pool-based active learning strategies to select informative data points from a large pool of unlabeled sentences (see section 4.1).

2. Data processing: the selected data is processed to remove redundancy when it exists such as duplicates and highly similar sentences, ensuring dataset diversity (see section 4.2).

3. Multi-round dataset preparation for FSL: multiple rounds of datasets are then created with incremental sample sizes for each active learning strategy (see section 4.3).

4. Model training: the PET model with mBERT is used to train and classify sentences as needing citations (see section 4.4).

Figure 1 provides a detailed flowchart of the ALPET model architecture, illustrating each of the steps mentioned above including an example of doing CWD with PET.

### 4.1. Data Selection

The CWD task is framed within a pool-based AL scenario using various query strategies. Central to these strategies is the acquisition function, a term commonly used in mathematical definitions to describe the mechanism that assigns a score to each unlabeled data point based on specific criteria. In the context of AL, this function is often referred to as a query framework or query strategy [16]. Traditionally, acquisition functions in AL are based on uncertainty or diversity, with more recent approaches incorporating hybrid methods that combine elements of both [91]. In this work, we employ uncertainty, diversity, and hybrid acquisition functions as part of our data subset selection process. Figure 2 illustrates the flowchart of data selection strategies. Next we elaborate these query strategies in more details.

### 4.1.1. Diversity sampling

The goal here is to select data points that are as different from each other as possible ensuring a representative sample of the overall data distribution. Three diversity sampling methods that we use in this work are geometry-based, corset-based, and cluster samplings.

A. **Geometry-based sampling.** Distance metrics belong in the geometry-based metrics, they play a fundamental role in identifying data points for labeling in the active learning scenarios. Among most widely used distance metrics are cosine and euclidean distance. We have used both of them in ALPET methodology and more details are presented below.

1. **Cosine distance** is derived from cosine similarity and measures dissimilarity between data points. In our task, sentences are represented by embeddings, or high-dimensional vectors, to capture the semantic content of the sentence. Cosine distance between two vectors $\mathbf{A}$ and $\mathbf{B}$ is mathematically defined as:

$$CosineDistance = 1 - \cos(\theta) = 1 - \frac{\mathbf{A} \cdot \mathbf{B}}{\|\mathbf{A}\| \times \|\mathbf{B}\|} \quad (1)$$

where $\mathbf{A} \cdot \mathbf{B}$ is the dot product of the vectors, and $\|\mathbf{A}\|$ and $\|\mathbf{B}\|$ are their magnitudes.

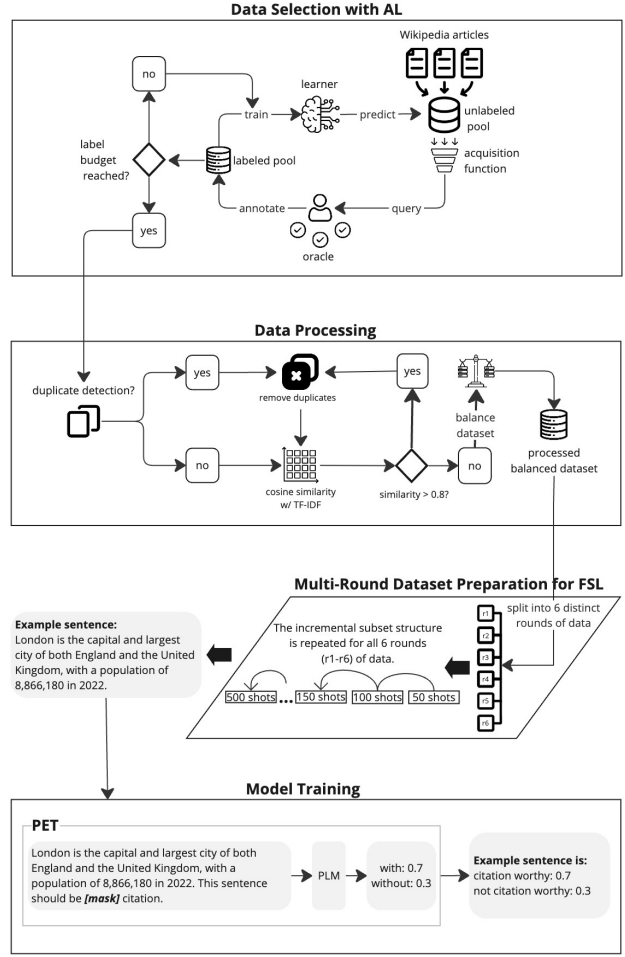

Figure 1: ALPET model architecture

Our goal in using cosine distance is to select sentences for labeling that represent different contexts where a citation might be needed. Sentences that are highly dissimilar from those already labeled might represent different styles, topics, or structures that the model has not yet encountered. We avoided using cosine similarity because we did not want to end up selecting sentences that are very similar to those already labeled as this could lead to redundancy in the training data, where the model continues to see variations of the same sentence structure or topic. In contrast, cosine distance encourages diversity in the selected data points, which we believe is crucial for effectively training the model in an active learning setting for the complex task of CWD.

2. **Euclidean distance** measures the straight-line distance between two points in a high-dimensional space. In our task, this metric is used to quantify the absolute difference between sentence embeddings, providing a direct measure of dissimilarity. The Euclidean distance between two vectors $\mathbf{A}$ and $\mathbf{B}$ is mathematically defined as:

$$EuclideanDistance = \|\mathbf{A} - \mathbf{B}\| = \sqrt{\sum_{i=1}^{n}(a_i - b_i)^2} \tag{2}$$

where $a_i$ and $b_i$ are the components of vectors $\mathbf{A}$ and $\mathbf{B}$ in an $n$-dimensional space.

Using euclidean distance, we aim to select sentences that are different from those already labeled, thus exploring underrepresented areas of the feature space. Unlike cosine distance, which focuses on the angular relationship between vectors, euclidean distance accounts for the overall magnitude of differences, making it particularly useful for detecting sentences that are not only contextually but also substantially different in their feature representations (such as length, complexity, or intensity). This diversity in selection might help minimizing redundancy and enhancing the model's ability to generalize across different types of citation-worthy content.

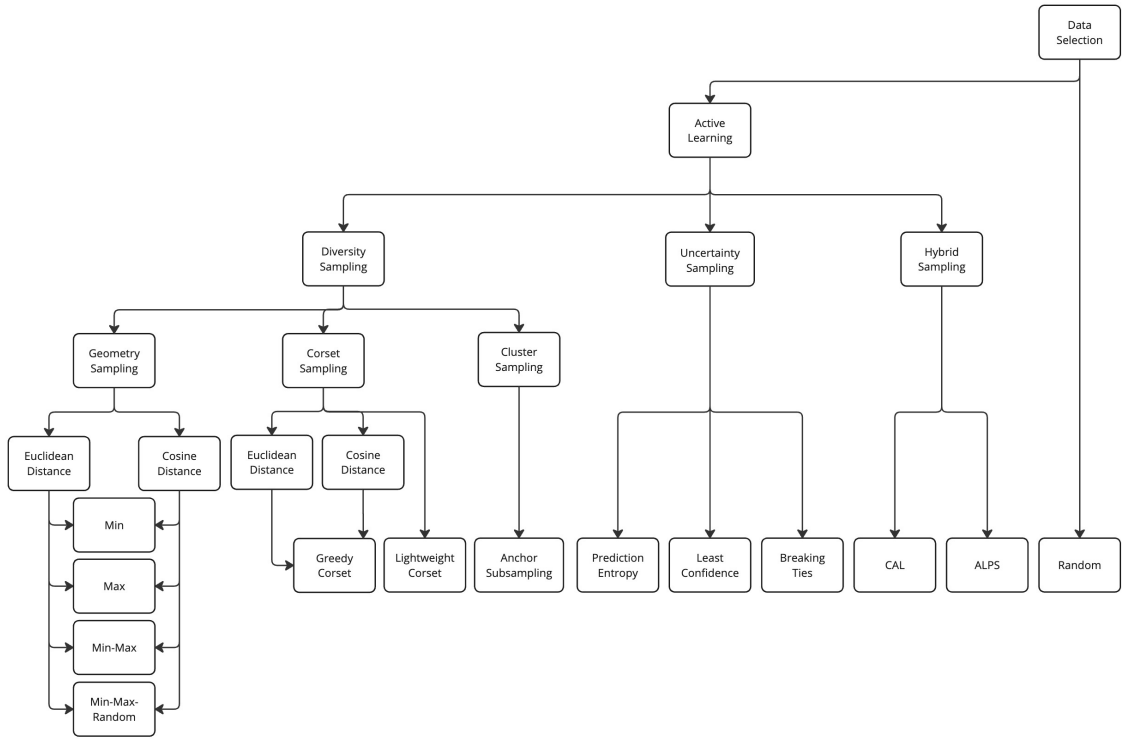

Figure 2: Data selection flowchart.

Building on distance metrics, in this research, we have used several custom data selection strategies to guide the active learning process. Each method leverages either cosine or euclidean distance to identify the most informative data points.

i. **Maximum average distance selection**: In this method, data points are selected based on the maximum average distance from already selected instances. The process begins with a cold start, where the first sentence is selected randomly from the unlabeled pool. For the next selection, the average distance of each candidate sentence (that has not been selected yet) to the already selected sentence is calculated. The sentence with the maximum average distance to the already selected sentence is chosen next. With two sentences selected, we recalculate the average distance of the remaining sentences to both selected sentences. The candidate with the highest average distance is added to the selection pool. This loop continues until the desired number of sentences is selected. By focusing on instances that are farthest from those previously selected, this approach aims to select points that are diverse and representative of different regions of the feature space.

ii. **Minimum average distance selection**: This method selects data points with the minimum average distance from the already selected instances. The selection process is similar to the one above that considers the maximum average distance, but instead, here we consider the minimum average distance. The rationale here is to focus on data points that are similar to those already chosen, effectively reinforcing the model's understanding of dense or well-represented areas of the feature space.

iii. **Combined maximum and minimum average distance selection**: This method alternates between selecting data points based on the maximum and minimum average distances from the already selected instances. The process begins with a cold start, where the first sentence is randomly chosen from the unlabeled pool. For the next selection, the average distance of each candidate sentence to the already selected sentence is calculated. The sentence with the maximum average distance to the selected instance is chosen first. Once this sentence is added to the selection pool, the next sentence is selected based on the minimum average distance from the already selected sentences. With three sentences now in the pool, the average distance of the remaining candidates to the selected sentences is recalculated. The selection alternates between choosing the sentence with the maximum average distance and the sentence with the minimum average distance, creating a balance between diversity and representativeness. This loop continues until the desired number of sentences is selected. By combining both maximum and minimum average distances, it ensures that both diverse and representative instances are included in each iteration, covering a broad spectrum of the data space while also reinforcing existing knowledge.

iv. **Maximum, minimum, and random selection**: In this method, we introduce an element of randomness. Each iteration involves selecting one data point with the maximum average distance, one with the minimum average distance, and one randomly chosen instance. This approach adds an exploratory component, allowing the model to occasionally consider unexpected instances that may not fit neatly into the established patterns, leading to accidental discoveries.

For each custom distance-metric selection strategy, the data points are initially selected from the unlabeled pool following the methodologies outlined above considering either cosine or euclidean distance. Once the selection process is complete, the selected data points are submitted to the oracle for labeling, with the labels being generated based on existing data. After the annotation has been finalized, the labeled data points are then input into the PET model to determine whether each sentence requires a citation.

B. **Corset-based strategies:** In AL a corset is a small, representative subset of the entire dataset that, when used to train a model, can approximate the performance of a model trained on the full dataset [92]. In our methodology we have used lightweight corset and greedy corset (with cosine and euclidean metrics).

i. **Greedy coreset**: Originally proposed to address the data labeling bottleneck for deep convolutional neural networks [92], this approach has been adapted for text data in the small-text library [93]. It constructs a greedy coreset over text embeddings, solving a k-center problem through a greedy approximation. The method aims for precise coverage and diversity by selecting points that minimize the maximum distance between any data point and its closest coreset point. While more computationally intensive, it is particularly useful when a highly representative subset is crucial, such as in smaller datasets.

ii. **Lightweight coreset**: This strategy selects a representative subset of data points using K-Means clustering, designed for computational efficiency by approximating the selection process [94]. It works by choosing data points that are farthest from the already selected points, minimizing redundancy. While this approach is efficient, it may not capture the full diversity of the data as precisely as more computationally intensive methods like GreedyCoreset [93]. The method's effectiveness depends on the quality of the feature representations, such as text embeddings.

C. **Clustering-based strategies:**

i. **Anchor subsampling:** This strategy addresses pool-based AL challenges of selecting minority class in large imbalanced datasets [95]. The term anchor here refers to the chosen class-specific instances from the labelled set. The process begins by selecting anchor points from the labeled dataset, representing different classes of the data space. Each unlabelled instance is then scored based on its average distance from the anchors, forming a subpool of the most similar instances. The next step is selecting the instances from the subpool to be labeled by an oracle. Once labeled, instances are added to the labeled dataset and the process repeats in subsequent AL iterations.

### *4.1.2. Uncertainty sampling*

Uncertainty-based selection focuses on identifying data points where the model is most unsure of its predictions.

i. **Prediction entropy:** This query strategy selects instances with the largest prediction entropy [93]. As a method it was initially proposed to reduce the labeling efforts in an object recognition task [96], however, in small-text library it was adapted for text-based data. Given a pool of unlabeled data, for each instance the model outputs a probability distribution over all possible classes. Then the entropy of the predicted probability distribution is calculated for each instance which are then ranked by the highest entropy scores indicating cases where the model is most uncertain about the prediction. The top k instances with the highest entropy scores are selected for labeling. Once labeled, instances are added to the training set, and the model is retrained on the updated labeled dataset.

ii. **Least confidence**: Is one of the earliest uncertainty-based strategies [97]. It selects instances with the least prediction confidence (regarding the most likely class) [93]. Specifically, for each instance in the unlabeled pool, the model assigns a probability distribution over the possible classes. Then it identifies instance where the model is least confident about the correct class. This instance is then selected for labeling because it represents a point of high uncertainty, where the model might benefit most from additional labeled data. However, this method may sometimes overlook instances where the model is uncertain between several classes because it only considers the confidence of the most likely class.

iii. **Breaking ties**: This function is designed to select data points which have a high uncertainty in classification, specifically those where the margin between the most likely and second most likely predicted class is minimal. The small margin indicates that the model is uncertain about which class the sentence belongs to, making these instances particularly valuable for AL. This strategy was originally proposed in the context of image data [98] but in small-text library it has been adapted to work with text data under the name BreakingTies [93]. The core idea to target ambiguous instances remains the same.

### *4.1.3. Hybrid sampling*

Hybrid acquisition functions were developed to combine the best aspects of both uncertainty and diversity sampling. In this research, we have used Contrastive Active Learning (CAL) and ALPS (Active Learning by Processing Surprisal), two state-of-the-art hybrid methods, to maximize the efficiency of the AL process.

i. **Contrastive active learning (CAL)**: This function combines uncertainty and diversity sampling for warm-start AL. Using ContrastiveActiveLearning from small-text library [93], we implement CAL [91], which fine-tunes BERT with an initial labeled dataset and applies a KNN algorithm to identify the closest labeled examples for each data point in the unlabeled pool. A contrastive score, based on Kullback-Leibler (KL) divergence between predicted probabilities of the unlabeled candidate and its labeled neighbors, is calculated to select high-divergence examples for labeling by a proxy. The labeled batch is then removed from unlabeled pool and added to the training (labeled) dataset. This loop repeats until all unlabeled data have been labeled. This method aims to effectively identify sentences with similar vocabulary but differing predictions, enhancing the selection of informative examples.

ii. **Active learning by processing surprisal (ALPS)**: This method combines uncertainty and diversity sampling for cold-start AL. Implemented as EmbeddingKMeans in small-text library [93], ALPS [99] uses surprisal embeddings derived from the masked language modeling loss in PLMs like BERT to estimate uncertainty, bypassing the need for unreliable model confidence scores in the cold-start scenario. After computing surprisal embeddings for each sentence in the unlabeled pool, K-Means is applied to cluster these embeddings and the sentence closest to each cluster center is selected. Thus ALPS identifies data points that are both surprising (indicating high uncertainty) and representative of diverse, underexplored areas in the data space, making it particularly effective in early stages where labeled data is scarce.

### *4.2. Data Processing*

#### *4.2.1. Duplicate and similarity removal*

In cold-start AL iterations, after each data subset selection, we observed duplicate sentences or sentences with high structural similarity. To mitigate redundancy in the dataset, we applied a cosine similarity with TF-IDF weights. Sentences with a cosine similarity score above 0.8 were considered too similar and were excluded from the final dataset. This threshold was selected based on our empirical analysis of the dataset, where we observed that sentences with a similarity score above 0.8 were almost identical except for minor variations, such as differing in only the last one or two words. For our CWD task, we aim to capture diverse sentence structures and contexts, so such highly similar sentences were unnecessary.

#### *4.2.2. Data balancing*

This research was conducted using Wikipedia articles where each was split into individual sentences, forming the unlabeled pool of data for AL. Although the sentences were pre-labeled, we temporarily removed the labels to simulate an active learning environment where labels are obtained iteratively from an oracle.

Each languages' dataset (ca, eu, and sq) is imbalanced, with higher proportion of sentences that do not contain citations compared to those that do. In pool-based AL scenarios, imbalance can become more pronounced due to the model's tendency to favor majority class examples, leading to a loop of oversampling the majority class. To address this, we applied random undersampling to the majority class post-selection, reducing its size to match the minority class. In this way the minority class remained sufficiently represented throughout the learning process which is critical for our task.

### *4.3. Multi-Round Dataset Preparation for Few-Shot Learning*

After data processing steps, each query strategy yielded a dataset with 3,000 data points per class. We then constructed six distinct rounds of datasets. Each round comprises ten subsets, with the number of data points per class varying from 50 to 500, increasing in increments of 50. The dataset preparation process involved the following steps:

1. **Round-based data partitioning**: We randomly separated the 3,000 data points per class into six distinct groups, each containing 1,000 total data points (500 per class), to allow for multiple experimental iterations. The six rounds were chosen in line with standard practices in machine learning experiments, where 5 to 10 rounds are typically used to ensure reliable and generalizable results.

2. **Incremental sample sizes**: For each group created in the first step, we generated ten cumulative subsets, with sample sizes ranging from 50 to 500 data points per class with increments of 50. This incremental approach is commonly used in few-shot learning experiments, as it allows for a fine-grained analysis of the model's performance across different levels of data availability.

### *4.4. Model Training with Pattern Exploit Training (PET)*

Pattern-Exploiting Training (PET) [100] is a semi-supervised method for few-shot learning in NLP tasks like text classification and natural language inference. The core idea behind PET is to reformulate input examples as cloze-style questions (fill-in-the-blank), that help PLMs better understand the task. This method uses a concept called Pattern-Verbalizer Pair (PVP) which includes two elements:

1. **Pattern** where the input is transformed into a fill-in-the-blank questions. This is done by inserting a masked token into the input text, which the model will later try to predict.

2. **Verbalizer** maps task labels to actual words in the language model's vocabulary. These words are what the model predicts to fill in the blank created by the pattern.

**Example:**

- **Input**: "This movie was amazing."

- **Pattern**: The input is transformed into "This movie was [mask]."

- **Verbalizer**: The words from model's vocabulary "great" (positive), "bad" (negative), and "okay" (neutral) are mapped to the task labels (positive, negative, neutral).

In this example, the model predicts the word that should replace *[mask]* based on the context of the sentence. The selected word is then compared to the verbalizer to determine the sentiment classification (e.g., if the model predicts "great," the sentiment is classified as positive).

## 5. Experimental Settings

### 5.1. Datasets

To validate our hypothesis, we use real-world data sourced from Wikipedia articles. Specifically, we employ three datasets in different languages: ca-citation-needed, eu-citation-needed, and sq-citation-needed [4] in Catalan, Basque and Albanian, respectively. These datasets contain contextual information beyond individual sentences and labels. For this study, however, we focused exclusively on two components: the text of sentences from Wikipedia articles and their labels indicating the presence of inline citations. The sentence text was used for data selection through various AL query strategies, while the labels served primarily to simulate the annotation process with oracles when necessary.

For each dataset — ca-citation-needed, eu-citation-needed, and sq-citation-needed — we applied a consistent labeling budget and data split, dividing the data into training, development, and testing sets.

In the context of FSL, where models are trained on very limited data, even slight changes in the test or development set sizes can sometimes affect performance [11]. Thus, in our experiments, we tested different sizes of the development and test sets, even though improvements in learning performance were not necessarily expected. The primary purpose of this approach was to assess how varying the size of these sets might influence the stability of our model's evaluation metrics. Although the stability and performance remained largely unchanged, we observed an increased in time and resource consumption when the test and development sets were used at their maximum capacity. Therefore, in our results, we report only the experiments where the number of shots in the test and development sets were limited. The data selection for these reduced sets was done randomly. Table 1 presents the details of the datasets used for the three languages, including their splits into training, development, and test sets. As described in section 4.3 for training PET models of each language we have created 6 distinct training datasets, each used to train a separate model. But we have evaluated each model using the same development and test datasets. The results of each language for all models per specific shots are then averaged and reported.

Table 1: Distribution of datasets and their usage across three languages. Numbers present the number of sentences (r1-r6 denote six rounds of distinct train sets).

| Dataset | Data Partition | Citation | No Citation |
|---|---|---|---|
| ca | | 335,538 | 802,052 |
| eu | Unlabeled Pool | 73,086 | 232,372 |
| sq | | 29,928 | 77,105 |
| ca/eu/sq | Train Sets r1-r6 | 500 each r1-r6 | 500 each r1-r6 |
| | Dev Set | 250 | 250 |
| | Test Set | 250 | 250 |

### 5.2. PET with Active Samples

Models that can be used with PET tasks are PLMs and in our CWD task we employed a multilingual BERT (mBERT) to calculate probabilities of candidate tokens that could replace *[mask]* in predefined patterns for each of the datasets we used. Since we are working with datasets of threee languages, we had to maually pre-define patterns for each language. In order to avoid introducing any bias in any of the languages we decided to use the same pattern structure for three languages but we translated them accordingly to match the language. Even though the goal of this research is not to find the most optimized patterns and verbalizer for PET, we experimented with a couple of patterns and we choose the best performing ones to report the final results on. It is worth mentioning that we started experimenting with patterns in Albanian language, then we translated them into Catalan which was quite straight forward. More challenging was translation of patterns into Basque language due to the grammar rules and structure of the language. We could not automatically just translate them with tools like Google Translate, instead we had to amend them manually in order for the patterns to make sense and to make sure that it follows grammatical rules of the language. The patterns we used for each language are presented in Table 2.

Table 2: Patterns used for PVP of PET model in three of the datasets ca-citation-needed, eu-citation-needed, sq-citation-needed.

| Language | Verbalizer | Pattern |
|---|---|---|
| ca | 0 : sense | 1. *[text_a]* Aquesta frase va *[mask]* citació. |
| | | 2. *[text_a]* Aquesta frase s'hauria d'escriure *[mask]* citació. |
| | | 3. *[text_a]* s'hauria d'escriure *[mask]* citar les fonts. |
| | 1: amb | 4. *[text_a]* En un article de Viquipèdia, aquesta frase sería *[mask]* citació. |
| | | 5. Si *[text_a]* fós part de Viquipèdia, els editors requeririen que s'escrigués *[mask]* citació per a que fosi verificable. |
| eu | 0 : gabe | 1. *[text_a]* Esaldi honetan erreferentzia *[mask]*. |
| | | 2. *[text_a]* Esaldi hau erreferentzia *[mask]* idatzi beharko litzateke. |
| | | 3. *[text_a]* idaztean erreferentzia *[mask]*. |
| | 1: barne | 4. *[text_a]* Wikipediako artikulu batean, esaldi honetan erreferentzia *[mask]*. |
| | | 5. *[text_a]* Wikipedian balego, egiaztagarritasuna mantentzeko editoreek gomendatuko lukete erreferentzia *[mask]* idaztea. |
| sq | 0 : pa | 1. *[text_a]* Kjo fjali duhet të jetë *[mask]* citim. |
| | | 2. *[text_a]* Kjo fjali duhet të shënohet *[mask]* citim.' |
| | | 3. *[text_a]* Duhet të shënohet *[mask]* burim informacioni. |
| | 1: me | 4. *[text_a]* Në një artikull të Wikipedias, kjo fjali duhet të jetë *[mask]* citim. |
| | | 5. Nëse *[text_a]* do të ishte fjali e Wikipedias, editorët do të kërkonin të ishte *[mask]* citim për shkak të verifikueshmërisë. |

### 5.3. Baseline Model

This study seeks to evaluate the efficiency of citation worthiness detection (CWD) within few-shot learning settings to accommodate languages with low-resources using the ALPET method.

We benchmark our work against the Contextualized Citation Worthiness (CCW) model [4] due to its SOTA approach in addressing CWD in the Wikipedia domain. The baseline incorporates a transformer-based architecture, utilizing contextual embeddings from mBERT to capture sentence-level features. To ensure methodological consistency and comparability, we adapted CCW by adding an AL step for data selection. This modification aligns with our ALPET approach, and it allow for a more direct performance comparison between the models in terms of both accuracy and efficiency across languages like Albanian, Basque, and Catalan.

### 5.4. Evaluation Metrics

To assess and compare the performance of the proposed ALPET method against the baseline CCW, we employ the macro F1 score as the primary evaluation metric. Given the balanced nature of our dataset and the equal importance of both classes—positive (sentences requiring inline citations) and negative (sentences not requiring inline citations)—the macro F1 score is well-suited for this task as it ensures that the performance across both classes is equally represented.

### 5.5. Training Details

**Hyperparameters**. The ALPET method was trained using the following hyperparameters. We utilized the multilingual BERT with a maximum sequence length of 256 tokens and a batch size of 4 for training and 8 for evaluation. The model was trained for 3 epochs, with a learning rate of 1e-5 and a weight decay of 0.01. Optimization was handled by the Adam optimizer with an epsilon value of 1e-8 and a maximum gradient norm of 1.0 to prevent gradient explosion. The same hyperparameters, where applicable, were used for the baseline CCW model. All experiments were executed on a GPU setup to handle the computational demands of training the BERT model.

**Checkpoints.** Training was repeated for 3 iterations to ensure robustness in the results. Checkpoints were saved after every epoch allowing the model to resume training from the last saved checkpoint.

**Labeling budget.** Our final labeling budget was 3000 samples per class for each AL query strategy. From which then we created six distinct rounds of datasets as described in section 4.3. For each dataset the maximum labeling budget was set to 500 instances per class, and we experimented with scenarios ranging from 50 to 500, increasing by steps of 50.

The task was conducted in a pool-based AL scenario, utilizing various AL query strategies as described in Section 4.1. AL labeling was done in iterations, with 60 samples selected per iteration across 100 iterations.

**Data split.** All labeled samples obtained through AL were used for training, while additional data, consisting of 250 instances per class, was reserved for test and development sets, as shown in Table 1.

**Hardware/Software setup.** All experiments were conducted using PyTorch on a single NVIDIA A100 40GB GPU.

## 6. Experiment Results

In this section we present the evaluation of our ALPET model alongside the CCW baseline model on three datasets CA-citation-needed, EU-citation-needed, and SQ-citation-needed. Results are organised as answers to hypotheses.

### 6.1. ALPET Outperforms Baseline Models in Low-Resource Languages (H1)

This section aims to evaluate Hypothesis *H1*. We hypothesized that ALPET would outperform the baseline CCW model in terms of data efficiency (i.e., achieving comparable performance with fewer labeled examples) and predictive performance (F1 Score) in low-resource languages, while utilizing the same AL query strategies.

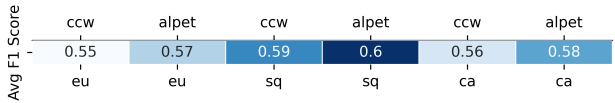

Figure 3: F1 Score averaged across all query strategies and instances count.

Our experiments conducted on the CA-citation-needed, EU-citation-needed, and SQ-citation-needed datasets provide empirical evidence that supports this hypothesis. Figure 3 offers a summarised view of the performance comparison between the ALPET model and the CCW baseline model. This figure presents the average F1 scores achieved by both models across all data selection strategies and instance counts for each of the three datasets. As figure indicates, ALPET, on average, achieves better performance than CCW baseline across all three datasets.

In the EU-citation-needed dataset as illustrated in the top subplot of Figure 4 (a), ALPET achieved an average F1 Score of 53% across all query strategies with only 50 labeled examples, whereas CCW required 150 labeled examples to reach the same F1 Score. This indicates that ALPET achieved comparable predictive performance with 66.67% reduction in labeled examples. In the SQ-citation-needed dataset, shown in the middle subplot of Figure 4 (a), ALPET achieved an average F1 Score of 58% across all query strategies with only 50 labeled examples, whereas CCW required 200 labeled examples to reach similar performance level. This represents a 75% reduction in labeled examples, reinforcing ALPET's superior data efficiency in this dataset.

Similarly, in the CA-citation-needed dataset, as presented in the bottom subplot of Figure 4 (a), ALPET achieved an average F1 Score of 55% across all query strategies with only 50 labeled examples, whereas CCW needed above 150 labeled examples to match this performance. Once again, this demonstrates that ALPET not only outperforms CCW in terms of predictive performance but also does so with at least 66.67% fewer labeled examples. The reduced annotation cost achieved by ALPET, showed by the smaller number of labeled examples required, is particularly valuable in low-resource settings where labeled data is scarce and expensive.

The heatmap in Figure 4 (b) provides a detailed comparison of AL query strategies, showing the average F1 scores across all instance counts for both ALPET and CCW in the three datasets (eu, sq, and ca). The heatmap demonstrates that ALPET generally outperforms CCW in most query strategies across all datasets. For example, in the pool-bt strategy, ALPET achieves an average F1 Score of 60% in the eu dataset, compared to 58% for CCW, and similarly outperforms in the sq and ca datasets with an average F1 Score difference of 2%-3%. However, the models achieve comparable performance for pool-cal and pool-greedyc in the sq dataset, and pool-greedyc-cosine for ca dataset. ALPET underperforms CCW in strategies such as cosine-max in the eu and sq datasets, euclidean-max-min in the eu dataset and euclidean-max in the sq dataset. In conclusion we can see that the results presented in this section demonstrate ALPET's ability to achieve comparable or higher F1 scores with significantly fewer labeled examples than the CCW baseline, supporting Hypothesis *H1*. ALPET shows a clear advantage in both data efficiency and performance for CWD in low-resource languages.

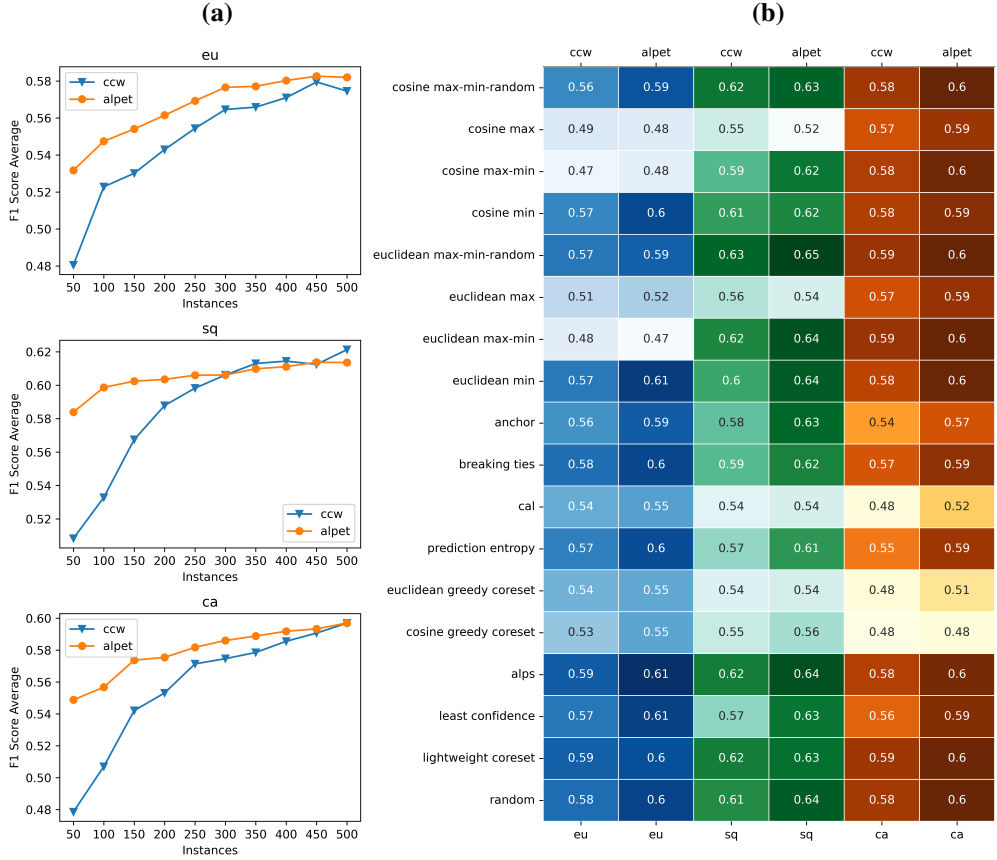

Figure 4: Comparison between the **ALPET** model and the **CCW** baseline. Subplots in figure (a) represents the average F1 Score for various active learning query strategies across different nr. of instances for three datasets: eu, sq, and ca. The orange line with circle markers represents the alpet model, while the blue line with triangle markers represents the ccw baseline. Each subplot in (a) corresponds to a specific dataset, with eu at the top, sq in the middle, and ca at the bottom. In figure (b), the heatmap displays the average F1 Score across all instances for each query strategy. The distinct colors in the heatmap (blue, green, and orange) correspond to the eu, sq, and ca datasets, respectively, while the x-axis distinguishes between the ccw and alpet models for each dataset.

### 6.2. ALPET's Performance Plateau and Data Efficiency (H2)

This section aims to evaluate Hypothesis *H2*. We hypothesized that ALPET's performance improves with increasing labeled data but plateaus after a certain number of samples, making it effective in low-resource settings. The subplots in Figure 5 confirm this hypothesis by showing that ALPET maintains a good performance up to 300 labeled examples, beyond which its improvement is minimal.

Looking closely at at the Figure 5a, at 50 shots, ALPET starts with a higher averaged F1 score across all three datasets (53%, 58%, and 55%) compared to CCW (48%, 51%, and 48%), showing a clear initial advantage in performance. As the number of labeled instances increases, ALPET continues to maintain its lead. The dashed vertical line

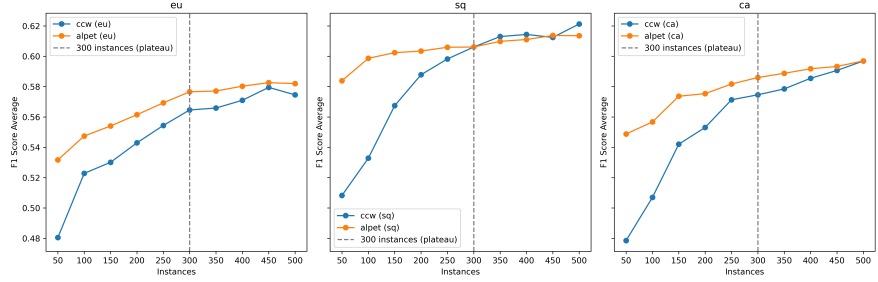

(a) F1 Score comparison between ALPET and CCW models across three datasets with plateau observed at 300 instances. Graphs show the average F1 score across all AL query strategies for ALPET and CCW models as the number of training instances increases.

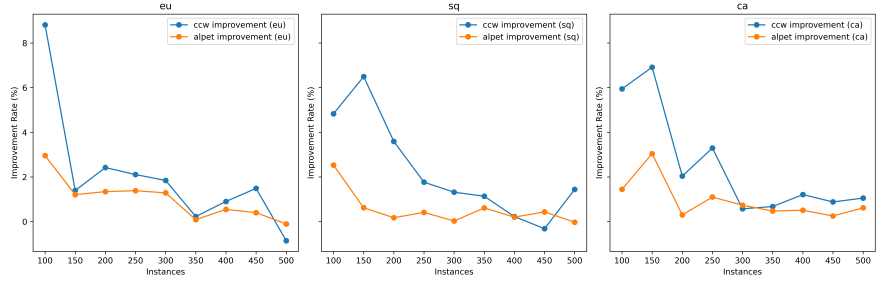

(b) Incremental F1 Score improvement between successive batches for ALPET vs. CCW across three datasets.

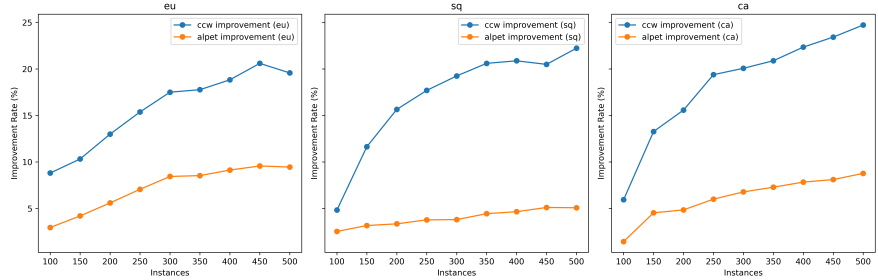

(c) Cumulative F1 Score improvement for ALPET vs. CCW across three datasets with varying numbers of training instances.

Figure 5: Comparison of F1 Score performance between ALPET and CCW models across three datasets (EU, SQ, CA). Subfigures show the average F1 scores, incremental improvements across batches, and cumulative F1 score improvements with increasing training instances.

marks the point where we hypothesized the performance plateau to occur (at 300 labeled examples per class). Specifically, in the EU-citation-needed dataset as shown in the Figure 5a (left plot), ALPET's performance rises from an F1 Score of 53% at 50 labeled examples to 58% at 300 labeled examples. Beyond 300 examples, the F1 Score stabilizes, showing minimal improvement with F1 score slightly passing 58% up to 500 instances. A similar trend is observed in the SQ-citation-needed dataset illustrated in the Figure 5a (middle plot), where ALPET's F1 score increases to 61% at 300 labeled examples, after which further improvement flattens. In the CA-citation-needed dataset the Figure 5a (right plot), ALPET reaches an F1 score of 59% at 300 labeled examples, beyond 300 instances and up to 500 the F1 Score improves to 60%. This trend is consistent across datasets, showing that ALPET is particularly efficient at leveraging smaller datasets in low-resource settings and that the extra effort or resources spent on labeling beyond this point does not lead to much further improvement. This is significant in low-resource settings, where obtaining large amounts of labelled data is often expensive and time-consuming. ALPET's ability to achieve good performance with only up to 300 labelled samples suggests its potential to overcome the data bottleneck and facilitate CWD in languages where labelled data is scarce.

A detailed comparison of performance gains in the Figure 5b shows the F1 score improvement between successive labeled example counts. For EU dataset Figure 5b (left plot) ALPET improves almost 3% from 50-100 instances, and 1-2% with each additional batch of labeled examples, but after 300 samples the gain diminishes to 0.1%. Similarly for CA datasets. An overall percentage improvement of instance count ranges is presented in the Table 3 which shows that from 300-500 instances the model improves about 1-2% across languages.

The subplots in the Figure 5c show the cumulative improvement in F1 Score relative to the initial performance at 50 instances for both ALPET and CCW models across the three datasets. ALPET shows steady improvement, with a maximum of 9% cumulative gain by 500 instances; with the majority of gain achieved up to 300 instances as seen in the column **100-300** of Table 3. In contrast, CCW achieves 25% gain by 500 instances reinforcing the need for more data to achieve comparable results.

ALPET's performance has direct implications for resource efficiency because it reduces the annotation effort required compared to approaches like CCW. Furthermore, training on a smaller dataset reduces the computational time and resources required, making ALPET a more efficient approach for CWD in environments with constrained resources.

The analysis presented in this section, demonstrating a clear performance plateau after 300 labeled examples across all three datasets, supporting Hypothesis *H2*. ALPET's ability to achieve and maintain good performance with a limited number of samples underscores its suitability for low-resource CWD tasks.

### 6.3. *Efficiency and Robustness of ALPET with Reduced Labeled Data in Low-Resource Languages (H3)*

This section aims to evaluate Hypothesis *H3*. We hypothesized that ALPET would achieve comparable performance to the baseline CCW model while requiring significantly fewer labeled examples. Specifically as presented in the Figure 5a, ALPET achieves an average F1 Score of 55% across the three datasets with 50 labeled examples per class, whereas CCW requires 200 examples for the same performance. We also evaluate whether ALPET's performance remains robust with fewer labeled examples to demonstrate its data efficiency in low-resource settings. The percentage reduction in labeled examples between ALPET and CCW for each query strategy was calculated by comparing the 50-shot performance of ALPET to the performance of CCW at the point where it matched ALPET's F1 Score. We consider an F1 Score of 55%, the average score achieved by ALPET with 50 labelled samples, as a competitive performance benchmark. A summary of reductions across three datasets is visualized in Figure 6 and presented in Table 4.

The reduction percentages across the three datasets reveal that ALPET achieves an average reduction of 70% in CA, 58% in EU, and 72% in SQ (mean in Table 4) while maintaining a competitive F1 Score of 55% compared to CCW performance. In at least 50% of the query strategies, ALPET reduces the labeled data requirement by 67% in CA and EU, and by 78% in SQ (median in Table 4). In 25% of the query strategies, ALPET achieves a reduction of 75-83% (75th percentile in Table 4). These results demonstrate that ALPET's reduction performance is consistently high across various query strategies. The variability (as indicated by std in Table 4) differs slightly across datasets. The EU dataset shows a standard deviation of 24%, suggesting higher variability in reduction percentages compared to SQ (21%) and CA (10%). This indicates that the reduction in labeled examples may depend more on the chosen query strategy for some languages than others.

Figure 6 visualizes the percentage reduction in labeled examples needed by ALPET compared to CCW across various query strategies for the CA, EU, and SQ datasets. Overall, ALPET demonstrates strong performance, with most query strategies achieving substantial reductions in labeling requirements, ranging between 58% and 83% across datasets. While the CA dataset shows consistently high reductions, the EU and SQ datasets exhibit some variability, including a few query strategies that fail to reduce labeling at all. However, the majority of strategies still perform effectively, highlighting the robustness of ALPET's approach in minimizing labeled data across different languages.

The findings presented in this section, demonstrating ALPET's ability to achieve competitive performance (F1 Score of 55%) with above 80% reductions in labelled data strongly support Hypothesis *H3*. ALPET's efficiency and robustness in low-resource settings make it a promising approach for CWD in such contexts. From the results above we saw that ALPET is efficient in three low-resource language, this might suggest that the model's data efficiency can generalize to other languages with similar resource constraints, however, further research will need to be conducted to confirm this.

### 6.4. *Effectiveness of Active Learning Query Strategies (H4)*

Initially we hypothesized that AL query strategies typically achieve higher F1 scores than random sampling in low-resource language datasets. This expectation was based on the belief that effective AL query strategies, which select data strategically, would outperform the simple random sampling. However, the results obtained from the comparative analysis of the ALPET model across three low-resource languages using the F1 Score as the evaluation metric did not support this hypothesis across all AL strategies that we employed.

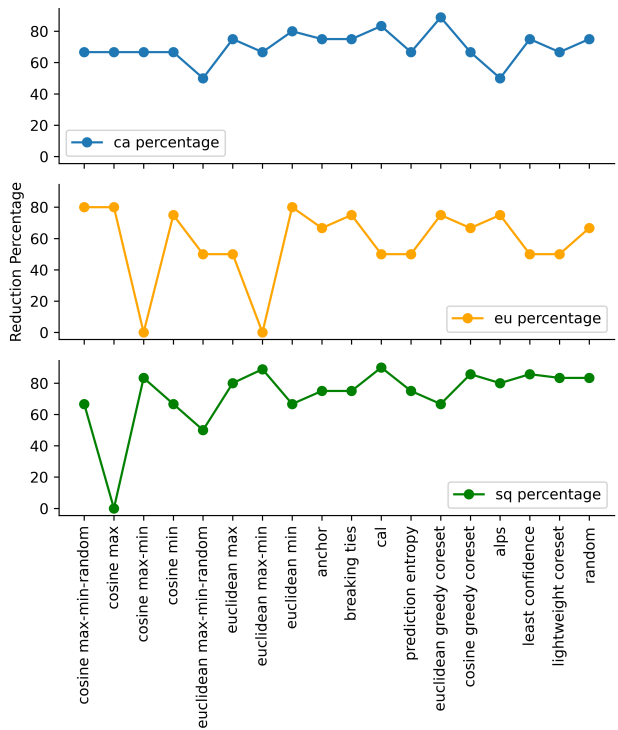

Table 3: Model percentage improvement by instance range count

| Dataset | Model | 50-100 | 100-300 | 300-500 |
|---------|-------|--------|---------|---------|
| EU | ALPET | 3.0% | 5.3% | 0.9% |
| SQ | ALPET | 2.5% | 1.2% | 1.2% |
| CA | ALPET | 1.4% | 5.3% | 1.9% |

Table 4: Summary of reduction percentages in labeled examples for ALPET compared to CCW

|  | Reduction (ca) | Reduction (eu) | Reduction (sq) |
|--------|--------|--------|--------|
| mean | 70% | 58% | 72% |
| std | 10% | 24% | 21% |
| median | 67% | 67% | 78% |
| 75p | 75% | 75% | 83% |

Figure 6: Percentage reduction in labeled examples needed by ALPET compared to CCW across query strategies for CA, EU, and SQ datasets.

As described in the dataset section, we created ten datasets for each language, with instances counts ranging from 50 to 500. In each case, we compared random sampling with one of the AL query strategies, ensuring that both methods used the same number of shots. In Figures 7 each dataset has its own plot (marked with ca, eu, sq at the top), whereas the fourth plot with notation (combined) holds the average performance of three datasets. The x-axes represent shot/instance sizes (50-500) while y-axis show AL strategies. To evaluate the performance, we calculated the difference in F1 scores between random sampling and AL strategies. Before generating the combined plot of Figure 7 we found the average F1 Score of three datasets and then performed the difference of F1 Score between random sampling and AL query strategies. We define an AL query strategy as better than random sampling if the difference between their F1 Scores is a negative number. The results were visualized using heatmaps, where red shades represent better performance by AL strategies relative to random sampling, the blue shades represent better performance from random sampling, and finally the light blue almost white squares with zero values represent similar performance of both models (see Figure 7).

The experiments conducted to evaluate this hypothesis revealed a more nuanced scenario than initially anticipated. While certain AL strategies did exhibit some advantages, the results only partially support H4. Specific AL strategies, such as ALPS, LightweightCoreset, euclidean-min, and euclidean-cycle outperform random sampling in specific instances. However, their effectiveness varied by language and shot size and often yielded only marginal improvements over random data selection. In contrast, AL strategies that failed to outperform random sampling, systematically across all shot sizes and languages, are cosine-max, euclidean-max, Greedy Coreset. More effective AL strategies, like ALPS and Lightweight Coreset, have in common usage of K-Means clustering which in essence works through euclidean distance calculation as does euclidean-min and euclidean-cycle.

In conclusion, the hypothesis that AL strategies consistently outperform random sampling in low-resource lan-guage datasets for CWD was not fully supported. The effectiveness of specific AL strategies was contingent upon factors such as the dataset, language, and the number of labelled instances. As a result, random sampling remains a competitive baseline, particularly when dealing with smaller pool of unlabeled data.

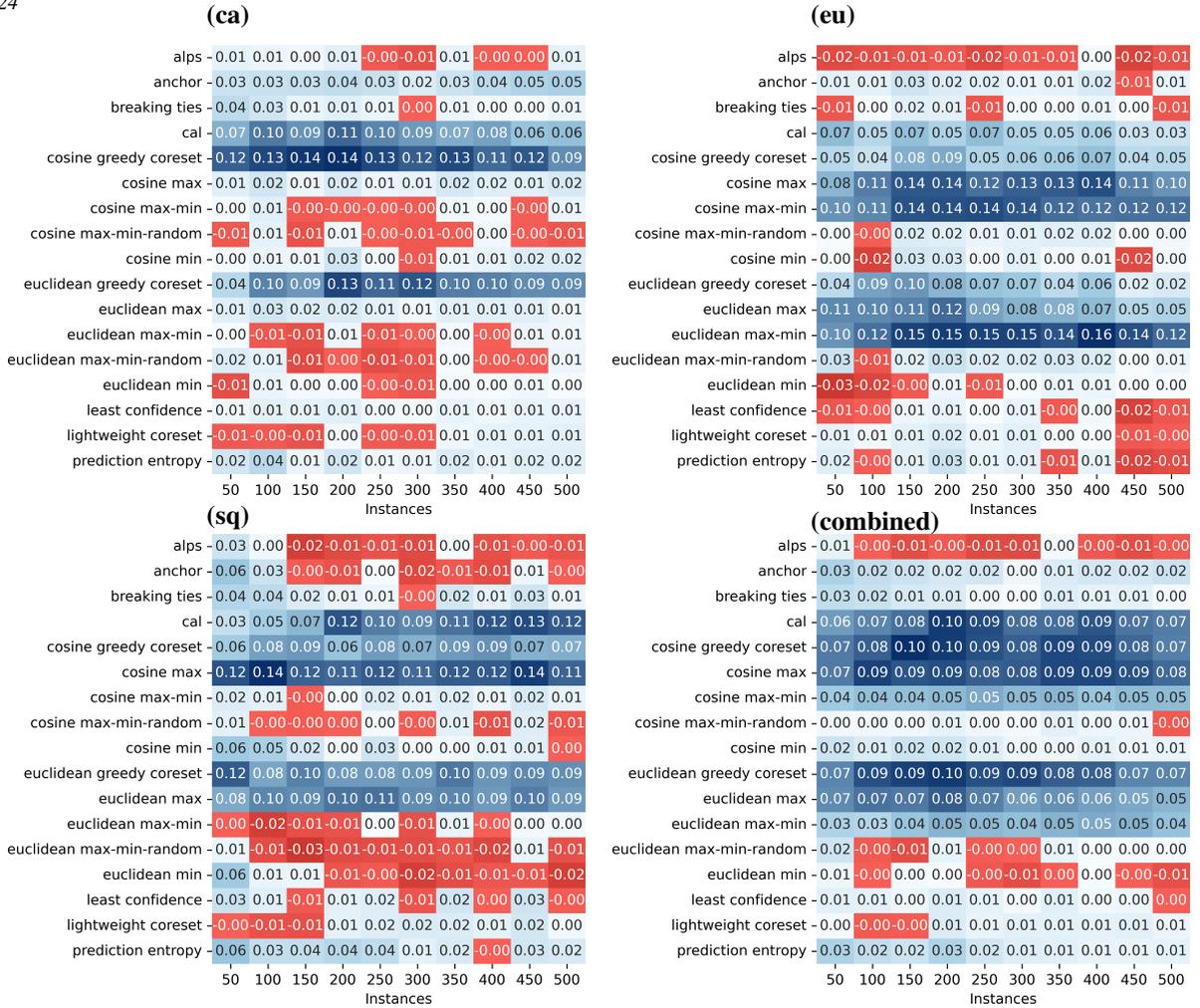

Figure 7: Comparing AL strategies to random data sampling. Heatmap displaying the cross-differences in F1 scores between active learning (AL) strategies and random sampling across all datasets and shot sizes.

### 6.4.1. Linguistic features and AL performance

The red squares in the heatmaps of Figure 7 show that performance of AL query strategies compared to random sampling varies across languages, which may be due to factors like language complexity. This variability signifies that further investigation is needed to understand the reason behind it.

In the ca and eu datasets, usually AL strategies outperform random sampling when the labeled data size is smaller (up to 250 shots) as opposed to when the instances increase up to 500. This may be attributed to the larger original unlabeled data pools in these two datasets, which give AL strategies access to a more diverse and varied set of examples to choose from. The higher variance in the data pool allows AL strategies to better identify informative samples, which leads to improved performance in low-shot scenarios. On the other hand, the sq dataset, with its smaller unlabelled data pool, shows more limited gains from AL strategies early on compared to when more labeled data instances are added. This can be an indication that AL's advantage is more pronounced when there is a larger pool of unlabeled data to choose from.

Ultimately, we were interested to find out in which cases AL query strategies are more consistent in beating random data selection across three languages. We found out that only in a limited number of cases this happened, ALPS with 250 and 300 instances and euclidean-min with 250 instances. Showing that the effectiveness of these strategies is more conditional than initially hypothesized. ALPS performance can be attributed to the strategy's ability to group semantically similar sentences based on dense embeddings from BERT, which encapsulates rich semantic information. By clustering these semantically dense sentences, K-Means increases the likelihood of selecting informative examples. In contrast, random sampling may pick redundant or less informative samples, making BERT embeddings and K-Means especially effective in low-resource settings.

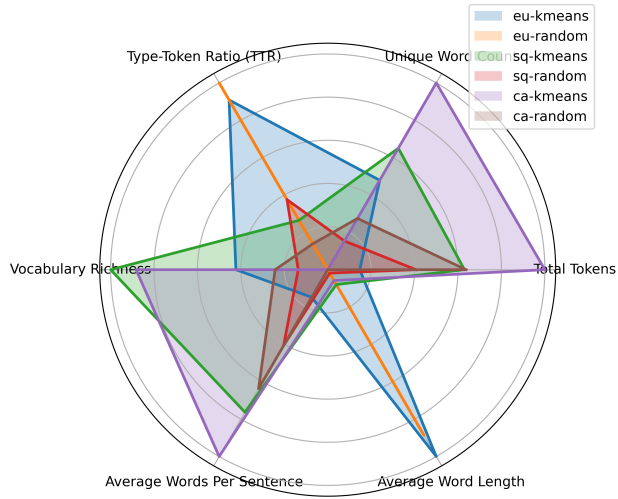

Figure 8: Linguistic features that could influence the performance of data selection through ALPS versus random selection.

To gain deeper insights into the linguistic features that influence the performance of AL strategies, we compared K-Means and random sampling across six linguistic metrics: Unique Word Count, Type-Token Ratio (TTR), Vocabulary Richness, Total Tokens, Average Words Per Sentence, and Average Word Length, visualized in Figure 8. The reduced TTR for K-Means suggest a trade-off between sentence length and lexical diversity because longer sentences often contain repeated words, which lower the TTR. However, the higher Unique Word Count and Vocabulary Richness indicate that the K-Means captures diverse vocabulary. By selecting longer sentences as presented by Average Words Per Sentence, K-Means samples provide more complex language structures. While TTR may be lower, the presence of more tokens suggests that the sentences are capable of conveying more information. K-Means, when paired with BERT embeddings, effectively captures semantically similar sentences. Average Word Length provides additional insight into linguistic complexity, particularly in Basque (eu), which, as an agglutinative language, naturally has longer words due to its morphological structure. K-Means' ability to work well with longer word forms in Basque suggests that it can effectively capture morphologically complex languages. In contrast, Albanian (sq) and Catalan (ca), which have shorter average word lengths, benefit more from K-Means' ability to select longer, more informative sentences rather than focusing on individual word length. In conclusion, K-Means performance across three languages indicates that this strategy as opposed to random has helped the model to better understand the language nuances that likely contribute to improved model training. This is crucial in low-resource contexts, where the richness of the training data influences the performance of models.

While not uniformly outperforming random sampling, analysis in this section reveal that specific AL query strategies, particularly those employing semantic clustering techniques like K-Means, can offer advantages in low-resource language settings. These advantages become more pronounced with larger unlabeled pool of data, where AL strategies have a greater opportunity to identify and leverage informative samples allowing random sampling to remain a competitive baseline for smaller datasets. These findings highlight the importance of carefully considering both the AL strategy and dataset characteristics when working with low-resource languages.

## 7. Conclusion

We have proposed ALPET as an active few-shot learning strategy to tackle the Citation Worthiness Detection (CWD) task for low-resource languages, integrating Active Learning (AL) and Pattern-Exploiting Training (PET). Our study investigates its effectiveness on three low-resource languages, i.e. Catalan, Basque, and Albanian. The resulting framework, ALPET, aimed to overcome the challenge of limited labeled data in these languages. Experiments on three datasets (CA-citation-needed, EU-citation-needed, and SQ-citation-needed) confirmed that ALPET outperforms the baseline CCW model in both data efficiency and accuracy (F1 Score).

One of the most significant findings is that ALPET requires significantly fewer labelled examples to achieve comparable or better performance than the CCW baseline. This is particularly valuable for low-resource languages, where labelled data is limited and expensive. With just 50 labelled examples per class, ALPET achieved an average F1 score of 55% across the three datasets. In contrast, CCW required 200 labelled examples for the same performance level. This demonstrates the remarkable data efficiency of ALPET. Furthermore, ALPET consistently achieved a considerable reduction in labelled data requirements while maintaining, or even surpassing, the baseline performance. Analysis revealed an average reduction of 70% for Catalan, 58% for Basque, and 72% in Albanian, highlighting the potential for above 80% reduction in annotation efforts in some of AL query strategies. In at least half of the query strategies, ALPET reduced the need for labelled data by 67% for Catalan and Basque, and by 78% for Albanian. This efficiency makes ALPET a very practical and cost-effective solution for CWD in low-resource settings. Another key finding that supports ALPET's suitability for low-resource scenarios is that its performance plateaus after 300 labelled samples, indicating its effectiveness even with smaller datasets. Figure 5a clearly illustrates this plateau, showing that beyond 300 labelled examples, further performance improvement is minimal across all datasets. This characteristic is particularly beneficial for low-resource languages, where obtaining extensive labelled datasets can be extremely challenging.

While the initial hypothesis suggested a general advantage for AL query strategies over random sampling, the experiments demonstrated a more complex scenario. Some AL strategies, particularly those using K-Means clustering, showed benefits, especially with larger pool of unlabeled datasets. However, their effectiveness was not universal and was often influenced by the dataset, language, and amount of labelled data. This suggests that the choice of AL strategy should be carefully considered based on the specific characteristics of the task and the available data.

The effectiveness of ALPET, especially its data efficiency and robustness in low-resource settings, suggests that the combined active few-shot learning approach holds great promise for CWD and has the potential to be extended to other NLP tasks facing similar constraints, such as claim detection or rumour detection. This research contributes to the development of effective CWD systems for under-resourced languages, paving the way for enhancing the reliability and trustworthiness of information in these languages, a crucial need in today's world where misinformation poses a significant threat. Future research could explore the impact of real-time human annotation in AL settings on CWD model performance, further refining and strengthening the applicability of this approach in real-world scenarios.

## 8. Limitation

A limitation of this study is the reliance on simulated AL rather than involving real-time human annotators. While AL is designed to iteratively select the most informative examples for human labeling to outperform random sampling, conducting such experiments with live human feedback is time-consuming and costly. This makes it challenging to implement in academic research. To address this, we simulated the process by using an already labeled datasets, treating them as if were unlabeled. Although this approach is widely adopted to bypass the logistical difficulties of real-time annotation, it may not fully replicate the dynamic interactions found in practical, human-in-the-loop AL environments. Human annotators bring their own subjective interpretations, biases, and inconsistencies to the labelling process, factors that a simulated environment cannot fully replicate.

Additionally, in this study we focused on a specific set of AL query strategies, while offering valuable insights it does not include the entire spectrum of available techniques. Exploring a wider range of AL methods, such as committee-based strategies, could provide a more comprehensive understanding of AL's potential and limitations for CWD in low-resource languages.

Future research could prioritise addressing these limitations by:

- Incorporating small-scale experiments with human annotators to validate the findings derived from the simulated AL setting. This would provide valuable insights into the real-world performance of ALPET and help refine the data selection process.

- Exploring and evaluating a more diverse set of AL strategies, focusing on their robustness and adaptability to different languages and data characteristics. This could involve investigating methods that specifically address the challenges posed by imbalanced datasets and noisy labels, which are common in low-resource scenarios.